\documentclass[conference]{IEEEtran}
\IEEEoverridecommandlockouts
\usepackage{cite}
\usepackage{amsmath,amssymb,amsfonts}
\usepackage{algorithmic}
\usepackage{graphicx}
\usepackage{textcomp}
\usepackage{xcolor}
\usepackage[hidelinks]{hyperref}

\usepackage{booktabs}
\usepackage{multirow}
\usepackage{orcidlink}
\usepackage{cleveref}
\usepackage{array}
\usepackage[caption=false]{subfig}
\usepackage{textcomp}
\usepackage{stfloats}
\usepackage{todonotes}

\def\BibTeX{{\rm B\kern-.05em{\sc i\kern-.025em b}\kern-.08em
    T\kern-.1667em\lower.7ex\hbox{E}\kern-.125emX}}
\begin{document}

\title{
    Whitening Consistently Improves Self-Supervised Learning
}


\author{
\IEEEauthorblockN{András Kalapos\,\orcidlink{0000-0002-9018-1372}, Bálint Gyires-Tóth\,\orcidlink{0000-0003-1059-9822}}
\IEEEauthorblockA{\textit{Department of Telecommunications and Artificial Intelligence, Faculty of Electrical Engineering and Informatics} \\
\textit{Budapest University of Technology and Economics}\\
Műegyetem rkp. 3., H-1111 Budapest, Hungary \\
\href{mailto:kalapos.andras@tmit.bme.hu}{kalapos.andras@tmit.bme.hu}
}
}

\maketitle

\begin{abstract}

Self-supervised learning (SSL) has been shown to be a powerful approach for learning visual representations. 
In this study, we propose incorporating ZCA whitening as the final layer of the encoder in self-supervised learning to enhance the quality of learned features by normalizing and decorrelating them.  
Although whitening has been utilized in SSL in previous works, its potential to universally improve any SSL model has not been explored. 
We demonstrate that adding whitening as the last layer of SSL pretrained encoders is independent of the self-supervised learning method and encoder architecture, thus it improves performance for a wide range of SSL methods across multiple encoder architectures and datasets.
Our experiments show that whitening is capable of improving linear and k-NN probing accuracy by 1-5\%. Additionally, we propose metrics that allow for a comprehensive analysis of the learned features, provide insights into the quality of the representations and help identify collapse patterns.
\end{abstract}


\section{Introduction}
Efficiently learning representations from unlabeled data has remained a fundamental challenge in machine learning since its inception. Learning on unlabeled data offers orders of magnitude larger scale datasets available for training larger and more capable models that exceed the capabilities of models trained via supervised learning. Self-supervised learning (SSL) is a key driver of recent advances in natural language processing foundation models and text-based multi-modal models through generative pretraining. The notion of foundation models emerged in part because these are shown to learn general knowledge through self-supervision that transfers well to many downstream tasks spanning a wider range than what was considered possible based on supervised pretraining. In computer vision, many SSL methods have been published, building upon predictive, instance discriminative or generative mechanisms. These methods have been shown to be effective in learning representations on large datasets and producing representations that transfer well to downstream vision tasks.

Normalization methods, such as Batch Normalization~\cite{BatchNormalizationAccelerating2015} and Layer Normalization~\cite{LayerNormalization2016} are key components of most deep neural networks. Ioffe and Szegedy~\cite{BatchNormalizationAccelerating2015} consider whitening as a technique to normalize and decorrelate activations, however, they discard it due to the computational cost and the challenges of backpropagation through the eigenvalue and eigenvector computation.
Recently multiple approaches have been published that utilize whitening in self-supervised learning, either as a mechanism to avoid representation collapse or to improve the quality of the learned features~\cite{WhiteningSelfSupervisedRepresentation2021,FeatureDecorrelationSelfSupervised2021,OvercomingDimensionalCollapse2024}. 
However, no comprehensive study has been published that investigates the effect of whitening on the performance of various self-supervised learning methods and encoder architectures. 

\textit{In this work} we introduce ZCA (Zero-phase Component Analysis) feature whitening in multiple self-supervised learning algorithms as the last layer of the encoder. In contrast to previous works, our approach of introducing whitening as the last layer of the encoder is orthogonal to the self-supervised learning method and encoder architecture, therefore provides a universal improvement to any self-supervised learning method. To support this claim, we conduct a comprehensive study on various self-supervised learning algorithms and encoder architectures.
We pretrain multiple encoders with different self-supervised learning methods and investigate the advantages of decorrelating encoder features via whitening. We analyze the learned representations through conventional linear and k-nearest neighbor probing and a set of metrics that allow for a comprehensive analysis of the learned feature space. 

\begin{figure}[t]
    \centering
    \includegraphics[width=\linewidth]{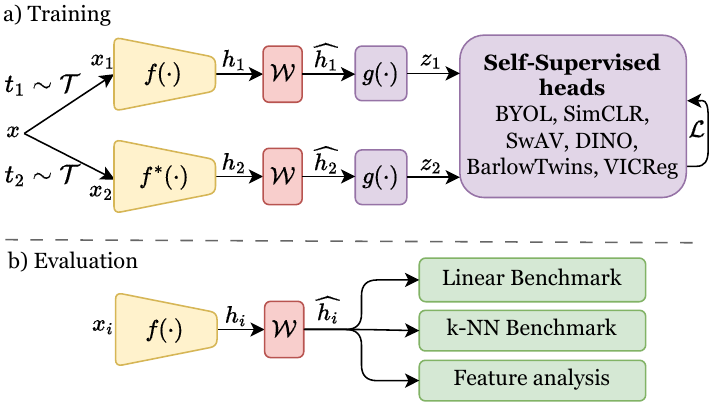}
    \caption{Overview of our method for decorrelating $h_i$ features via ZCA whitening, denoted by $\mathcal{W}$. $\widehat{h_i}$ represent whitened features. We apply feature whitening on both branches during pretraining using all self-supervised learning methods that we investigate. Whitening is also applied during evaluation.} 
    \label{fig:arch}
  \end{figure}
 
The main contributions of the current work are the following:
\begin{itemize}
  \item It is demonstrated that introducing whitening as the last layer of the encoder consistently improves the performance of multiple self-supervised learning methods and encoder architectures. The orthogonality of this whitening approach to the self-supervised learning algorithm and encoder architecture are shown via a comprehensive study. The proposed approach can provide 1-5\% linear and k-NN probing accuracy gains on CIFAR10, STL10 and Tiny-ImageNet datasets. 
  \item A framework is proposed for analyzing learned features through human-interpretable simple metrics or descriptors that allow for a comprehensive analysis of the learned feature space. 
\end{itemize}

Source code supporting the findings of this study is available at \url{https://github.com/kaland313/SSL-Whitening}.

\section{Related work}
Self-supervised learning is a field of machine learning that aims to learn representations from unlabeled data. It is becoming a dominant paradigm in natural language processing (NLP) and speech technologies and is also heavily researched in computer vision. However, the differences in the domains are significant, text data represents much higher-level concepts and is more structured, while visual data has a much lower semantic level and is often higher dimensional. Self-supervised learning relies on pretext tasks, which are more intuitive, and empirically better performing in NLP, while in vision the pretext tasks need to encourage learning from low-level features to high-level semantics. 
  
Over the past years, many methods have been proposed for visual self-supervised learning. Most of these build on instance discrimination learning, where the model is trained to distinguish between different views of the same image. This is done by creating different augmented versions of the same image (called views) and training the model to predict if the views are from the same image or not. Primarily, loss function formulation and exact training objectives differentiate these approaches from each other. Contrastive or deep metric learning methods (e.g.~\cite{SimpleFrameworkContrastive2020}) rely on similarity measures of embeddings, learning to minimize it between positive pairs and maximize it between negative pairs. Positive pairs are views from the same image, while negative pairs are pairs of views from different images. Self-distillation methods (e.g.~\cite{BootstrapYourOwn2020,EmergingPropertiesSelfSupervised2021}) learn to predict the representation corresponding to one view from a different view of the same image via a predictor network. Other methods~\cite{UnsupervisedLearningVisual2020,BarlowTwinsSelfSupervised2021,VICRegVarianceInvarianceCovarianceRegularization2022,WhiteningSelfSupervisedRepresentation2021} formulate the learning objective based on clustering or correlation analysis of the representations. Masked image modelling methods~\cite{MaskedAutoencodersAre2022,DesigningBERTConvolutional2022} are based on principles closely related to the masked language modelling pretext task of NLP and show good representation learning capabilities in vision as well.

Decorrelated Batch Normalization (DBN) based on ZCA whitening was proposed by Huang~et~al.~\cite{DecorrelatedBatchNormalization2018}, serving as an alternative to Batch Normalization that not only normalizes the data but also decorrelates it. They derive gradient computation for whitening and suggest conducting whitening on feature groups for efficiency. Additionally, they highlight the instability of PCA whitening due to axis swapping and introduce ZCA whitening as a solution. 
In a follow-up work, Huang~et~al.~\cite{IterativeNormalizationStandardization2019} presents an effective whitening approach based on Newton's iteration. 
As one of the first works applying whitening to self-supervised learning, Ermolov~et~al.~\cite{WhiteningSelfSupervisedRepresentation2021} introduces W-MSE, a self-supervised learning (SSL) method utilizing whitening to prevent representation collapse. 
Hua~et~al.~\cite{FeatureDecorrelationSelfSupervised2021} proposes to use DBN in self-supervised learning, replacing the Batch Normalization layer in the projector, resulting in enhanced representation quality and avoidance of feature collapse. Finally, Hassanpour~et~al.~\cite{OvercomingDimensionalCollapse2024} introduces whitening through Iterative Normalization to enhance feature quality in the MoCo-V2 self-supervised learning method~\cite{ImprovedBaselinesMomentum2020}. They replace the last Batch Normalization layer in the encoder and demonstrate performance enhancements in medical image segmentation tasks.

\section{Methods}
\subsection{Feature whitening in self-supervised learning}

Whitening decorrelates feature dimensions and avoids complete or dimensional collapse by simple linear transformations of the feature space. We propose to introduce ZCA (Zero-phase Component Analysis) whitening transformation as the last layer of the encoder in self-supervised learning methods to improve the quality of the learned features (see \cref{fig:arch}).


ZCA whitening is a linear transformation that scales and rotates the feature space to make the features uncorrelated and have unit variance. 
Given a matrix $\mathbf{X} \in \mathbb{R}^{n \times f}$ (where $n$ is the number of samples and $f$ is the number of features) the whitening transformation first centers the features by subtracting the means $\mathbf{X_c}=
\left(\mathbf{X}-\mu \cdot \mathbf{1}^T\right)$, where $\mu$ is the mean of the features and $\mathbf{1}$ is a vector of ones. Then we compute the covariance matrix $\mathbf{C} = \frac{1}{n}\mathbf{X_c}^T\mathbf{X_c}$ and its eigenvalue decomposition $\mathbf{C} = \mathbf{D}\mathbf{\Lambda}\mathbf{D}^T$. 
ZCA whitening then transforms a matrix $\mathbf{X}$ into $\mathbf{X}_{ZCA}$ by the following formula:

\begin{equation}\label{eq:zca}
  \mathbf{X}_{\text{ZCA}}=\mathbf{D} \mathbf{\Lambda}^{-\frac{1}{2}} \mathbf{D}^T\left(\mathbf{X}-\mu \cdot \mathbf{1}^T\right)
\end{equation}

Implementing ZCA whitening based on \cref{eq:zca} requires the computation of the covariance matrix and its eigenvalue decomposition, which is computationally expensive. Therefore, we implement feature whitening by Iterative Normalization~\cite{IterativeNormalizationStandardization2019} as a differentiable and efficient method based on Newton's iteration. As pretraining is performed on minibatches, we whiten features on a minibatch level both during training and evaluation. Batch-wise whitening was shown to be effective by Huang~et~al.~\cite{DecorrelatedBatchNormalization2018}. In many cases, large batch size requires group-wise whitening to avoid excessive memory consumption of the covariance matrix. However, we use a small batch size in our experiments, therefore we use batch-wise whitening. 

A growing body of literature~\cite{DecorrelatedBatchNormalization2018,IterativeNormalizationStandardization2019,WhiteningSelfSupervisedRepresentation2021,FeatureDecorrelationSelfSupervised2021,OvercomingDimensionalCollapse2024} points to the benefit of feature decorrelation and mechanisms avoiding dimensional collapse in self-supervised learning. Intuitively correlated features are redundant leaving room for more information to be captured by the feature space. Similarly, dimensional collapse leads to a feature space that captures less information than a fully utilized one, leading to lower-quality representations. Therefore, techniques that perform feature space whitening, such as Decorrelated Batch Normalization~\cite{DecorrelatedBatchNormalization2018} and Iterative Normalization~\cite{IterativeNormalizationStandardization2019} have the potential to improve the quality of the learned features. 
The insertion of a whitening layer to the end of the encoder is orthogonal to the self-supervised learning method and encoder architecture therefore it can improve any self-supervised pretraining approach. We apply the whitening during training and evaluation as well.


\subsection{Enhanced evaluation method for feature space analysis}\label{sec:metrics}
We propose a set of metrics that allows for a comprehensive analysis of the learned feature space. Specifically, we selected metrics that measure human-interpretable properties of embeddings and can give insight into the performance they achieve. Dimensional collapse is an important, yet hard-to-detect issue in self-supervised learning that can lead to poor performance of some algorithms, therefore we choose metrics that can show such representation collapse.

\subsubsection{Mean absolute feature correlation ($\overline{R_{ij|i\neq j}}$)}
Hua~et~al.~\cite{FeatureDecorrelationSelfSupervised2021} proposes the mean absolute correlation of features as a metric to identify feature collapse. They argue that a strong correlation between the feature axis is associated with collapsed dimensions. It is computed as the mean of the absolute value of off-diagonal elements of the correlation matrix $\textbf{R}$:

\begin{equation}
  \overline{R_{ij|i\neq j}} = \frac{1}{f\cdot (f-1)}\sum_{ij|i\neq j} R_{ij}, \mathrm{where}\ 
  R_{i j}=\frac{C_{i j}}{\sqrt{C_{i i} * C_{j j}}}
\end{equation}

where $\textbf{C}$ is the covariance matrix of the feature matrix $\textbf{H}$, while $C_{i i}$ and $C_{i j}$ correspond to variance and covariance terms in it. $\textbf{H} \in \mathbb{R}^{n\times f} $, where $n$ is the number of samples, $f$ is the number of output features of the encoder. 

While Hua~et~al.~\cite{FeatureDecorrelationSelfSupervised2021} compute this metric on the projected features ($z$ on \cref{fig:arch}), we compute it directly on encoder features ($\widehat{h_i}$ on \cref{fig:arch}). Although this doesn't lead to as trivially interpretable and clean results, it is more informative, directly measures the encoders' properties and reveals the degree of how related, redundant or collapsed the features are.

\subsubsection{Mean feature standard deviation ($\overline{s^*}$)}
Similarly to feature correlation, Hua~et~al.~\cite{FeatureDecorrelationSelfSupervised2021} proposes the mean standard deviation of features as a metric that can reveal collapse patterns, complete or dimensional collapse. It is computed as the mean of (corrected) sample standard deviation of feature dimensions in the $\textbf{H}$ feature matrix.
\begin{equation}
  \overline{s^*} = \frac{1}{f} \sum_{j}^{f} s^*_j = \frac{1}{f} \sum_{j}^{f} \sqrt{\frac{1}{n-1} \sum_{i}^{n} (H_{ij} - \overline{H_j})^2}
\end{equation}


\subsubsection{Anisotropy}
Following research from natural-language-processing~\cite{AnisotropyTrulyHarmful2023,ShapeLearningAnisotropy2023} and self-supervised learning on event sequences~\cite{SelfSupervisedLearningEvent2024}, we propose the anisotropy of the feature matrix as a metric of self-supervised representation quality. Though not in complete consensus, previous works show that lower values of anisotropy correspond to higher quality representations and higher performance. In the context of our research, we hypothesize that low anisotropy can be linked to a diverse feature space that captures as many details as possible and cannot be projected to a lower-dimensional manifold. Conversely, high anisotropy results from a quickly decaying singular value spectrum, which is indicative of a feature space that can be collapsed onto a low-dimensional subspace.  Anisotropy is computed as the ratio of the squared first (largest) singular value over the sum of squared singular values: 
\begin{equation}
\operatorname{Anisotropy}(\textbf{H})=\frac{\sigma_1^2}{\sum_i \sigma_i^2}
\end{equation}

where $\sigma_i$ are (ordered) singular values of the \textbf{H} feature matrix.

\section{Experiments}

\subsection{Data}\label{sec:data}
We conduct experiments on the CIFAR10\cite{LearningMultipleLayers2009}, STL10\cite{AnalysisSingleLayerNetworks} and Tiny-ImageNet\cite{TinyImagenetVisual2015} datasets, for quick experimentation and to provide a clear comparison between the different self-supervised learning methods. 
For self-supervised pretraining, linear and k-NN probe fitting we use the standard training set of the CIFAR10 and compute all evaluation metrics on the standard test set. In the case of STL10, we use the training and unlabeled partitions for self-supervised pretraining, use the training (labelled) set for fitting the probe layers and report all evaluation metrics on the test set. On Tiny-ImageNet, we use the standard training and validations sets for pretraining, probe training and evaluation respectively.

We use each dataset on its original resolution and only apply resizing and cropping to generate views for the self-supervised learning methods. Normalization is performed using the mean and standard deviation statistics derived from ImageNet, maintaining consistency with established practices in the literature. For data augmentation, we adhere to the specific settings published with each self-supervised learning method employed. However, an exception is made for SwAV and DINO, which rely on multi-cropping strategies. In these cases, we adjust the local cropping strategy to better suit the lower resolution of the datasets and omit the blur augmentation entirely. 

\subsection{Neural Network architecture}\label{sec:network_arch}


In this study, we investigate ResNet\cite{DeepResidualLearning2016} and ConvNeXtV2\cite{ConvNeXtV2Codesigning2023} encoders. The widespread adoption of ResNets in SSL literature provides a clear motivation for utilizing this architecture. Additionally, most SSL methods that we investigate were developed on ResNet-50, making hyperparameter selection simpler and grounded in literature. We opt for the smaller ResNet-18 variant to better fit the datasets that we use. As a modern convolutional network that rivals the performance of vision transformers~\cite{ImageWorth16x162020}, 
ConvNeXtV2~\cite{ConvNeXtV2Codesigning2023} was developed specifically for self-supervised pretraining. Additionally, ConvNeXts feature multiple design choices that differentiate them from ResNets, such as Layer Normalization~\cite{LayerNormalization2016} and depth-wise separable convolutions. This makes them ideal for testing the generalizability of SSL methods to less commonly used architectures.

Self-supervised learning methods require additional layers such as predictors or projector networks. We use these as published in original papers with minor adjustments to the dimensions of these layers to better suit the smaller resolution and complexity of the datasets used in this study. In the case of Barlow Twins, we use a two-layer projector network instead of the three-layer one used in the original paper to better match the size of the projector to the datasets. This is based on CIFAR10 experiments by Chen~and~He~\cite{ExploringSimpleSiamese2021}. 



\subsection{Self-supervised learning methods}\label{sec:self_supervised_learning_methods}
We investigate the following self-supervised learning methods: BYOL, VICReg, SimCLR, DINO, SwAV, Barlow Twins~\cite{BootstrapYourOwn2020,VICRegVarianceInvarianceCovarianceRegularization2022,SimpleFrameworkContrastive2020,EmergingPropertiesSelfSupervised2021,UnsupervisedLearningVisual2020,BarlowTwinsSelfSupervised2021}. These methods are chosen due to their widespread usage and proven performance across various tasks. Furthermore, they have publicly available implementations, facilitating easy experimentation and comparison. Importantly, they represent a diverse array of self-supervised learning principles and loss functions, offering insights into the effectiveness of different approaches in representation learning. 

\begin{figure*}[tb]
    \centering
    \includegraphics[width=\linewidth]{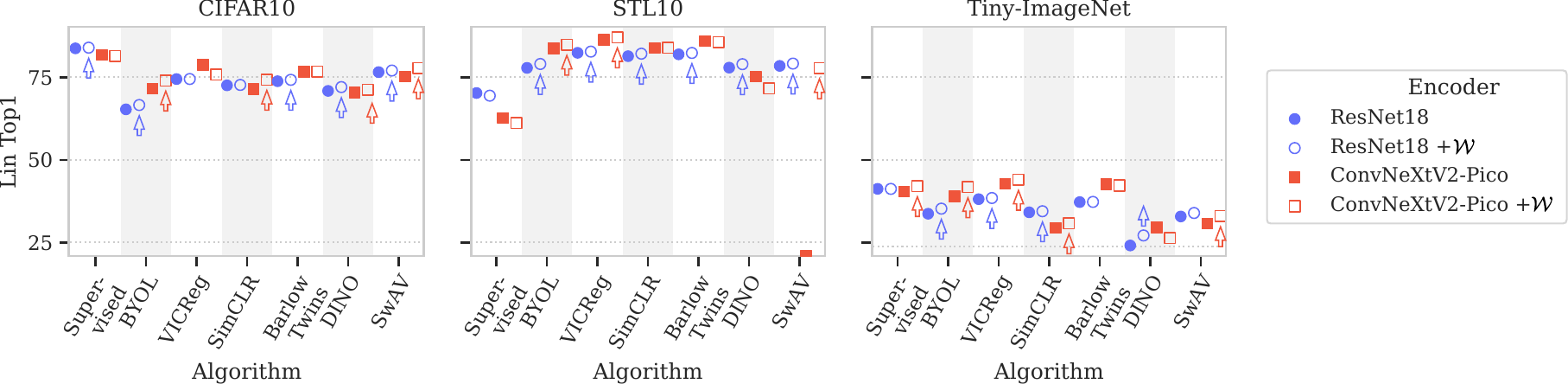}
    \caption{Linear probe results for all self-supervised learning methods and encoders with and without whitening. $\mathcal{W}$ indicates whitening as the last layer of the encoder. Arrows indicate where whitening provides a notable improvement in accuracy. (For ResNet18 the markers show mean accuracy over 5 runs.)}
    \label{fig:main_results}
\end{figure*}

\subsection{Evaluation}\label{sec:evaluation}
Following common practice in SSL literature, our primary objectives are linear and k-nearest neighbor probing accuracy. We employ both metrics to assess the quality of the representations learned by each encoder. We compute these using frozen encoders on normalized images, fit the linear and k-NN classification heads on the training partitions and report top-1 and top-5 accuracies on the test sets. 


In addition to the common probe accuracy scores, we investigate the quality of the learned features along the metrics presented in \cref{sec:metrics}. All metrics are computed based on a validation feature matrix, which is acquired with minimal augmentations, such as normalization and deterministic cropping.


\subsection{Training, hyperparameters and implementation details}
We aim to run all self-supervised learning methods with the same hyperparameters, to ensure a fair comparison. 
However, it is challenging to find a hyperparameter set that fits all methods. We empirically find that BYOL required a lower learning rate than the other methods, the rest of the hyperparameters are the same for all methods.
For all methods, we use the AdamW~\cite{DecoupledWeightDecay2018} optimizer, with a learning rate of 0.001, weight decay coefficient of 
$10^{-6}$ and a batch size of 256. We train for 200 epochs on all datasets, with a constant learning rate during the training. For BYOL, the learning rate is 0.0001. 

We run all experiments on a GPU cluster, each experiment using one NVIDIA A100 40 GB GPU with 16 CPU cores (AMD EPYC 7763) and 64 GB RAM. Training times for 200 epochs are 0.5-3 hours on CIFAR10 and 1-16 hours on STL10 and Tiny-ImageNet. 

The cumulative runtime for all experiments in this research is 1600 hours with an estimated total emission of 165 kgCO$_2$eq\footnote{Estimations were conducted using the \href{https://mlco2.github.io/impact}{MachineLearning Impact calculator} presented in~\cite{lacoste2019quantifying}}.

We implement all experiments based on the Lightly~\cite{Lightly2020} self-supervised learning library and Pytorch Lightning~\cite{PyTorchLightning2019a}. ResNet and ConvNeXt architectures are implemented in the Pytorch Image Models~\cite{PyTorchImageModels2019a} library.

\section{Results}

\subsection{Linear and k-NN probing accuracy}
\Cref{fig:main_results} presents linear probe accuracy for all self-supervised learning methods and encoders with and without using Iterative Normalization-based whitening. As highlighted by the arrows indicating improvement, whitening provides better or comparable performance to the baselines. The improvement is consistent across multiple self-supervised learning methods for both ResNet ConvNeXt encoders. Supervised pretraining presents an exception, showing a slight decrease in performance in most cases. 
Another notable outlier is the result on the STL10 dataset, ConvNeXtV2-Pico encoder with the SwAV method, where the training collapses without whitening, while whitening prevents this collapse, allowing SwAV to perform on par with the other methods.

\begin{figure}[tb]
  \centering
  \includegraphics[width=\linewidth]{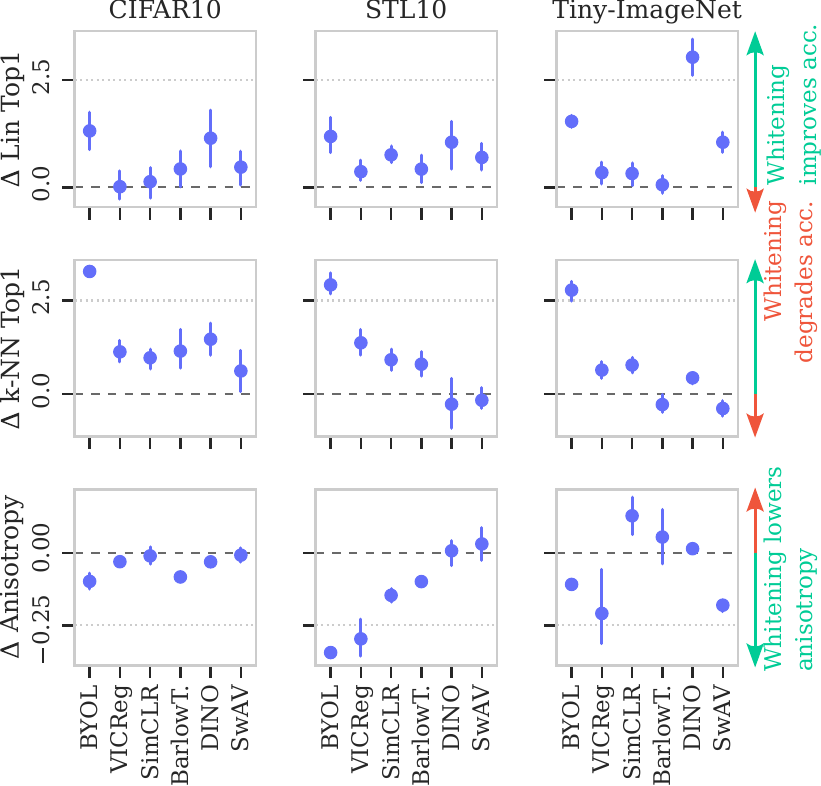}
  \caption{Probe accuracy gains and feature anisotropy change provided by the addition of whitening to a ResNet18 encoder with 5 different random seeds. Dots show mean accuracy gain over 5 runs, error bars show the 95\% confidence intervals. Positive probe values indicate an improvement in accuracy, negative values indicate a decrease in accuracy. Whitening lowers the anisotropy of features, indicating more decorrelated features. 
  }
  \label{fig:probe_diff}
\end{figure}

Additionally, we report the improvement in probe accuracies with confidence intervals over runs with 5 different random seeds using the ResNet-18 encoder (see \cref{fig:probe_diff}). The figure shows that the improvement is robust to random initialization and measurable in both linear separability and k-NN separability of the features. Among all methods, whitening improves the performance of BYOL the most on all three datasets, with $>1\%$ improvement in linear probe accuracy and $>3\%$ in k-NN probe accuracy. For other methods, both probe accuracies show mostly improvements or no significant performance degradations on all three datasets. 

\newcommand{\multirowrotate}[2]{\multirow[c]{#1}{*}{\rotatebox[origin=c]{90}{#2}}}

\subsection{Convergence and Compute efficiency}

\begin{figure}[tb]
  \centering
  \includegraphics[width=\linewidth]{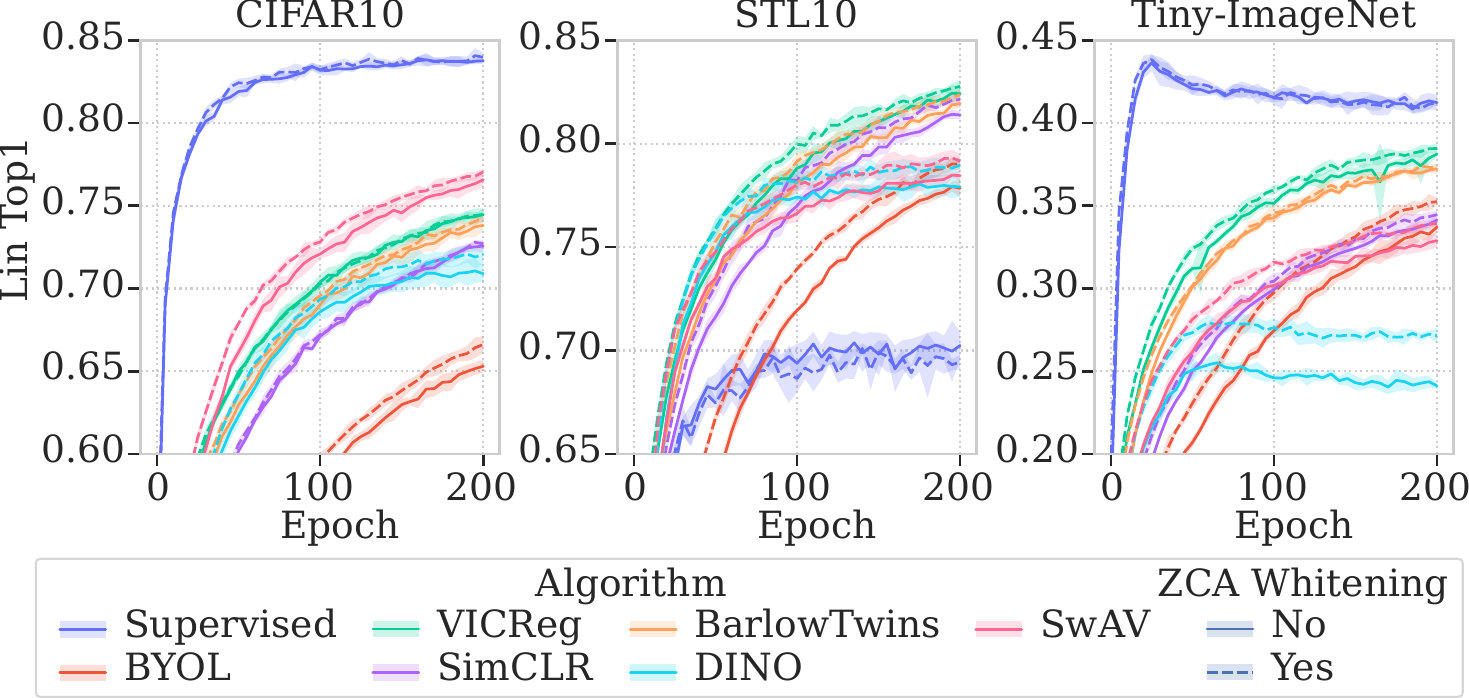}
  \caption{Learning curves for pretraining a ResNet-18 encoder with and without whitening. The curves show the mean linear probe accuracy over 5 runs, while shaded bands correspond to its standard deviation.}
  \label{fig:learning_curves}
\end{figure}

\Cref{fig:learning_curves} presents the learning curves for pretraining a ResNet-18 encoder with and without whitening. It shows that whitening not only improves the final accuracy but improves the accuracy consistently during training, essentially lowering the required number of epochs to reach the same accuracy. This is especially visible for BYOL, DINO and SwAV.

The additional computation time required for whitening increases overall training times by 15 minutes on average for CIFAR10 and STL10, while adding less than 25 minutes to the training time on Tiny-ImageNet. More efficient implementations of whitening could improve this overhead further. 

\subsection{Analysis of the learned features}\label{sec:analysing-learned-features}

\begin{figure}[tb]
  \centering
  \includegraphics[width=\linewidth]{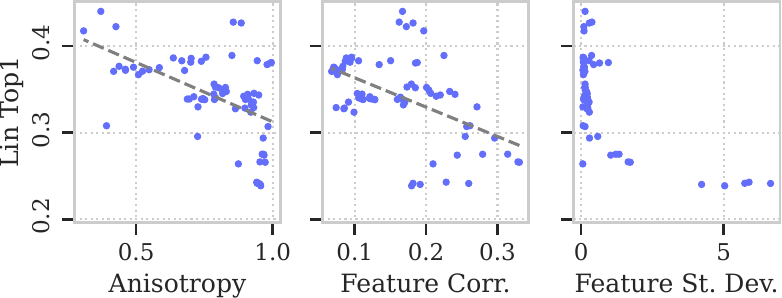}
  \caption{Scatter plots of feature metrics against linear probe accuracy on the Tiny-ImageNet dataset for all self-supervised learning methods and encoders, with and without whitening.}
  \label{fig:feature_metrics}
\end{figure}

In this section, we analyze the learned features based on quantitative measures that describe distributions and statistical properties of the features.
  
\Cref{fig:probe_diff} shows that improvements in k-NN accuracy provided by whitening are consistent with lower feature anisotropy for most self-supervised learning methods presented in the figure. This indicates that whitening provides improved feature quality that can be also measured through the anisotropy of the feature space. Similar trends for the effect of whitening on feature correlation, variance and rank are not as clear, therefore we don't include these in the figure. 

To further investigate the utility of the proposed metrics to provide insight into the quality of the learned features we present scatter plots of each metric against the linear probe accuracy in \cref{fig:feature_metrics}. 
The figure shows that anisotropy and feature correlation metrics are correlated with linear probe accuracy, with lower metric values corresponding to higher probe accuracy. 
High feature standard deviation is consistent with low linear probe accuracy, however for low standard deviation values the relation with probe accuracy is less clear. 
For models that show various representation collapse patterns, these metrics can provide useful insight, as dimensional and complete collapse results in lower variance and rank. 

Results for anisotropy and feature correlation on STL10 and Tiny-ImageNet are aligned with intuitive expectations. More correlated features are redundant and capture less information therefore providing lower probe accuracy.
High anisotropy corresponds to a feature space that has low $\sigma_{i|i\neq1}$ singular values compared to $\sigma_1$ first singular value and represents features that lie in a lower-dimensional subspace of the embedding space. This effectively lower dimensional feature space can capture less information, hence leading to lower classification accuracy.

\section{Conclusion}
This work investigates the impact of Zero-phase Component Analysis (ZCA) whitening on the performance of self-supervised learning (SSL) methods. We introduce whitening as the last layer of the encoder in various SSL algorithms and demonstrate its effectiveness in improving the quality of the learned features. We analyze the quality of the learned features with a variety of metrics, including linear and k-NN probing accuracy and quantitative analysis of the learned feature space. Our experiments on CIFAR10, STL10, and Tiny-ImageNet datasets show that whitening improves linear and k-NN probe accuracy by 0.5-3\% and 1-5\% respectively.  
Additionally, we propose a set of metrics that allows for a comprehensive analysis of the learned feature space, including ones that can indicate different collapse patterns. Experiments on multiple datasets and pretraining approaches support that these metrics correlate with probe accuracy, providing a useful tool for analyzing the learned representations. 

\section*{Acknowledgements}
ChatGPT‑4o was used to improve the quality and phrasing of the text of this paper.

We acknowledge KIFÜ (Governmental Agency for IT Development, Hungary, \url{https://ror.org/01s0v4q65}) for awarding us access to the Komondor HPC facility based in Hungary.

\bibliographystyle{IEEEtran}
\bibliography{IEEEabrv,references}

\clearpage
\section*{Supplementary Material}

\subsection{Model sizes}
We provide a table for the number of trainable parameters for each encoder and self-supervised head in \cref{tab:model_parameters}.

\begin{table}[hb]
    \centering
    \caption{Summary of the model parameters for the encoders and self-supervised heads.}
    \label{tab:model_parameters}
    \begin{tabular}{llc}
    \toprule
    & Model & \# Parameters \\
    \midrule
    \multirow[c]{2}{*}{Encoder} & ResNet-18 & 11.2 M \\
    & ConvNeXtV2-Pico & 8.55 M \\
    \midrule
    \multirow[c]{6}{*}{SSL Head} & BYOL head (Projector, Predictor) & 1.32 M \\
    & SimCLR (Projector) & 1.32 M \\
    & VICReg (Projector) & 5.25 M \\
    & Barlow Twins (Projector) & 5.25 M \\
    & DINO (Projector) & 0.69 M \\
    & SwAV (Projector, Prototypes) & 0.39 M \\
    \bottomrule
    \end{tabular}
\end{table}

\subsection{Detailed results}
\Cref{table:detailed_results} shows detailed results of linear and k-NN probe accuracy, as well as the accuracy gain provided by the whitening. The results are consistent with those of \cref{fig:main_results}, showing a 0.5-5\% increase in accuracy scores. One notable outlier is the result on the STL10 dataset, ConvNeXtV2-Pico encoder with the SwAV method, where the training collapses without whitening, while whitening prevents this collapse. In this case, whitening provides a 62\% increase in linear probe accuracy and a 56\% increase in k-NN probe accuracy, allowing SwAV to perform on par with the other methods.

\makeatletter
    \setlength\@fptop{0\p@}
\makeatother

\begin{table}[t!]
  \centering
  \caption{Detailed results on three datasets. $\mathrm{\Delta}$ shows the improvements provided by whitening. Positive values indicate improvement and are highlighted in bold. Observe that most reported gains are positive. BL and $+\mathcal{W}$ indicate the accuracy without (baseline) and with whitening, respectively.
  }
  \label{table:detailed_results}
  \begin{tabular}{ccrcccccc}
  \toprule
  \multirowrotate{3}{Dataset} & \multirowrotate{3}{Encoder} & {} & \multicolumn{3}{c}{Lin Top1} & \multicolumn{3}{c}{k-NN Top1} \\
  {} & {} & {} &  {} & {} & {} & {} & {} & {} \\
  {} & {} & {Algo.} & {BL} & {$+\mathcal{W}$} & {$\mathrm{\Delta}$} & {BL} & {$+\mathcal{W}$} & {$\mathrm{\Delta}$} \\
  \midrule
  \multirowrotate{14}{CIFAR10} & \multirowrotate{7}{ResNet18} & \itshape \color{gray}Superv. & \itshape \color{gray} 83.76 & \itshape \color{gray} 83.98 & \bfseries \itshape \color{gray} 0.21 & \itshape \color{gray} 83.38 & \itshape \color{gray} 83.72 & \bfseries \itshape \color{gray} 0.34 \\
   &  & BYOL & 65.30 & 66.61 & \bfseries 1.31 & 59.80 & 63.07 & \bfseries 3.28 \\
   &  & VICReg & 74.47 & 74.48 & \bfseries 0.01 & 71.67 & 72.79 & \bfseries 1.13 \\
   &  & SimCLR & 72.56 & 72.69 & \bfseries 0.13 & 69.72 & 70.69 & \bfseries 0.97 \\
   &  & BarlowT & 73.81 & 74.24 & \bfseries 0.43 & 70.96 & 72.11 & \bfseries 1.15 \\
   &  & DINO & 70.89 & 72.04 & \bfseries 1.14 & 67.29 & 68.75 & \bfseries 1.46 \\
   &  & SwAV & 76.58 & 77.05 & \bfseries 0.47 & 73.95 & 74.56 & \bfseries 0.61 \\
  \cmidrule{2-9}
  & \multirowrotate{7}{ConvNeXtV2-Pico} & \itshape \color{gray}Superv. & \itshape \color{gray} 81.79 & \itshape \color{gray} 81.47 & \itshape \color{gray} -0.32 & \itshape \color{gray} 81.74 & \itshape \color{gray} 81.55 & \itshape \color{gray} -0.19 \\
   &  & BYOL & 71.57 & 74.04 & \bfseries 2.47 & 66.85 & 71.10 & \bfseries 4.25 \\
   &  & VICReg & 78.78 & 75.82 & -2.96 & 75.40 & 73.56 & -1.84 \\
   &  & SimCLR & 71.45 & 74.27 & \bfseries 2.82 & 72.27 & 73.59 & \bfseries 1.32 \\
   &  & BarlowT & 76.71 & 76.70 & -0.01 & 73.00 & 73.64 & \bfseries 0.64 \\
   &  & DINO & 70.37 & 71.25 & \bfseries 0.88 & 64.07 & 67.86 & \bfseries 3.79 \\
   &  & SwAV & 75.24 & 77.74 & \bfseries 2.50 & 74.96 & 76.00 & \bfseries 1.04 \\
  \midrule
  \multirowrotate{14}{STL10} & \multirowrotate{7}{ResNet18} & \itshape \color{gray}Superv. & \itshape \color{gray} 70.25 & \itshape \color{gray} 69.44 & \itshape \color{gray} -0.81 & \itshape \color{gray} 69.85 & \itshape \color{gray} 69.93 & \bfseries \itshape \color{gray} 0.07 \\
   &  & BYOL & 77.85 & 79.03 & \bfseries 1.19 & 73.11 & 76.02 & \bfseries 2.92 \\
   &  & VICReg & 82.41 & 82.78 & \bfseries 0.36 & 79.21 & 80.58 & \bfseries 1.37 \\
   &  & SimCLR & 81.39 & 82.15 & \bfseries 0.75 & 78.42 & 79.33 & \bfseries 0.91 \\
   &  & BarlowT & 81.95 & 82.37 & \bfseries 0.42 & 79.15 & 79.94 & \bfseries 0.80 \\
   &  & DINO & 77.93 & 78.98 & \bfseries 1.05 & 75.04 & 74.76 & -0.28 \\
   &  & SwAV & 78.46 & 79.15 & \bfseries 0.69 & 75.24 & 75.07 & -0.17 \\
  \cmidrule{2-9}
  & \multirowrotate{7}{ConvNeXtV2-Pico} & \itshape \color{gray}Superv. & \itshape \color{gray} 62.70 & \itshape \color{gray} 61.11 & \itshape \color{gray} -1.59 & \itshape \color{gray} 63.04 & \itshape \color{gray} 62.03 & \itshape \color{gray} -1.01 \\
   &  & BYOL & 83.73 & 84.86 & \bfseries 1.14 & 76.58 & 81.85 & \bfseries 5.27 \\
   &  & VICReg & 86.29 & 87.11 & \bfseries 0.82 & 84.88 & 85.06 & \bfseries 0.19 \\
   &  & SimCLR & 83.93 & 83.94 & \bfseries 0.01 & 80.06 & 81.51 & \bfseries 1.45 \\
   &  & BarlowT & 85.95 & 85.63 & -0.33 & 83.76 & 83.73 & -0.04 \\
   &  & DINO & 75.18 & 71.63 & -3.55 & 70.88 & 68.46 & -2.41 \\
   &  & SwAV & 13.84 & 77.74 & \bfseries 63.90 & 17.01 & 74.54 & \bfseries 57.53 \\
  \midrule
  \multirowrotate{14}{Tiny-ImageNet} & \multirowrotate{7}{ResNet18} & \itshape \color{gray}Superv. & \itshape \color{gray} 41.24 & \itshape \color{gray} 41.23 & \itshape \color{gray} -0.01 & \itshape \color{gray} 40.24 & \itshape \color{gray} 41.85 & \bfseries \itshape \color{gray} 1.61 \\
   &  & BYOL & 33.72 & 35.25 & \bfseries 1.54 & 26.77 & 29.55 & \bfseries 2.78 \\
   &  & VICReg & 38.13 & 38.47 & \bfseries 0.34 & 32.29 & 32.92 & \bfseries 0.64 \\
   &  & SimCLR & 34.15 & 34.47 & \bfseries 0.32 & 27.38 & 28.16 & \bfseries 0.77 \\
   &  & BarlowT & 37.24 & 37.30 & \bfseries 0.06 & 31.58 & 31.29 & -0.29 \\
   &  & DINO & 24.11 & 27.14 & \bfseries 3.03 & 19.92 & 20.35 & \bfseries 0.43 \\
   &  & SwAV & 32.89 & 33.94 & \bfseries 1.05 & 24.65 & 24.26 & -0.39 \\
  \cmidrule{2-9}
  & \multirowrotate{7}{ConvNeXtV2-Pico} & \itshape \color{gray}Superv. & \itshape \color{gray} 40.47 & \itshape \color{gray} 42.10 & \bfseries \itshape \color{gray} 1.63 & \itshape \color{gray} 41.86 & \itshape \color{gray} 42.40 & \bfseries \itshape \color{gray} 0.54 \\
   &  & BYOL & 38.92 & 41.76 & \bfseries 2.84 & 31.79 & 34.48 & \bfseries 2.69 \\
   &  & VICReg & 42.77 & 44.01 & \bfseries 1.24 & 36.71 & 37.65 & \bfseries 0.94 \\
   &  & SimCLR & 29.39 & 30.81 & \bfseries 1.42 & 27.85 & 27.75 & -0.10 \\
   &  & BarlowT & 42.67 & 42.25 & -0.42 & 36.21 & 35.89 & -0.32 \\
   &  & DINO & 29.58 & 26.41 & -3.17 & 21.89 & 21.08 & -0.81 \\
   &  & SwAV & 30.72 & 32.99 & \bfseries 2.27 & 26.17 & 25.76 & -0.41 \\
  \bottomrule
  \end{tabular}
\end{table}

\end{document}